\documentclass[conference]{IEEEtran}
\IEEEoverridecommandlockouts
\usepackage{cite}
\usepackage{amsmath,amssymb,amsfonts}
\usepackage{algorithmic}
\usepackage{graphicx}
\usepackage{textcomp}
\usepackage{xcolor}
\usepackage{graphicx}
\usepackage{subcaption}  
\usepackage{tabularx}
\usepackage{booktabs}    
\usepackage{tabularx}    
\usepackage{times}
\usepackage{helvet}
\usepackage{multirow}
\usepackage{subcaption}
\usepackage{amssymb}
\usepackage{amsfonts}
\usepackage{amsmath}
\usepackage{courier}
\usepackage{threeparttable}
\usepackage{xcolor}
\usepackage[font=small]{caption}
\usepackage{url}
\usepackage{amssymb}
\usepackage{algorithm}
\usepackage{algorithmic}
\usepackage{svg}
\usepackage{newfloat}
\usepackage{listings}
\def\BibTeX{{\rm B\kern-.05em{\sc i\kern-.025em b}\kern-.08em
    T\kern-.1667em\lower.7ex\hbox{E}\kern-.125emX}}
\begin{document}

\title{RASR: Retrieval-Augmented Super Resolution for Practical Reference-based Image Restoration}

\author{
    \IEEEauthorblockN{Jiaqi Yan$^{1,2*}$\thanks{* This work was done during the author's internship at TeleAI.}, Shuning Xu$^{2,3}$, Xiangyu Chen$^{2\dagger}$\thanks{$\dagger$ Corresponding Author.}, Dell Zhang$^2$, Jiantao Zhou$^{3}$, Jie Tang$^1$, Gangshan Wu$^1$, Jie Liu$^{1\dagger}$}
    \IEEEauthorblockA{$^1$ State Key Laboratory for Novel Software Technology, Nanjing University}
    \IEEEauthorblockA{$^2$ Institute of Artificial Intelligence (TeleAI), China Telecom}
    \IEEEauthorblockA{$^3$ State Key Laboratory of Internet of Things for Smart City,}
    \IEEEauthorblockA{Department of Computer Science and Information Science, University of Macau}
    \IEEEauthorblockA{jiaqi$\_$yan@smail.nju.edu.cn, chxy95@gmail.com, liujie@nju.edu.cn}}
\maketitle

\begin{abstract}
Reference-based Super Resolution (RefSR) improves upon Single Image Super Resolution (SISR) by leveraging high-quality reference images to enhance texture fidelity and visual realism. However, a critical limitation of existing RefSR approaches is their reliance on manually curated target-reference image pairs, which severely constrains their practicality in real-world scenarios. 
To overcome this, we introduce Retrieval-Augmented Super Resolution (RASR), a new and practical RefSR paradigm that automatically retrieves semantically relevant high-resolution images from a reference database given only a low-quality input. This enables scalable and flexible RefSR in realistic use cases, such as enhancing mobile photos taken in environments like zoos or museums, where category-specific reference data (e.g., animals, artworks) can be readily collected or pre-curated. 
To facilitate research in this direction, we construct RASR-Flickr30, the first benchmark dataset designed for RASR. Unlike prior datasets with fixed target-reference pairs, RASR-Flickr30 provides per-category reference databases to support open-world retrieval.
We further propose RASRNet, a strong baseline that combines a semantic reference retriever with a diffusion-based RefSR generator. It retrieves relevant references based on semantic similarity and employs a diffusion-based generator enhanced with semantic conditioning.
Experiments on RASR-Flickr30 demonstrate that RASRNet consistently improves over SISR baselines, achieving +0.38 dB PSNR and -0.0131 LPIPS, while generating more realistic textures. These findings highlight retrieval augmentation as a promising direction to bridge the gap between academic RefSR research and real-world applicability.
\end{abstract}


\begin{figure*}[t]
  \centering
  \includegraphics[width=\textwidth,keepaspectratio]{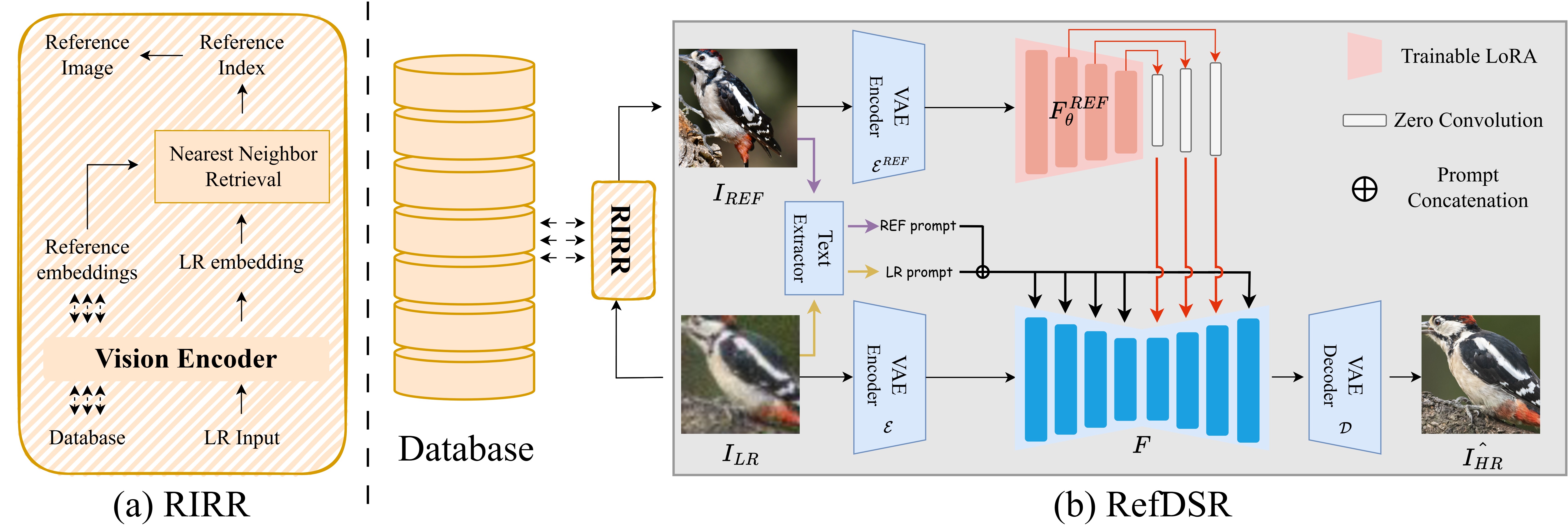}
  \caption{Overall architecture of RASRNet. It consists of two modules: Real-world Image Reference Retriever (RIRR) and Reference Diffusion Super Resolution (RefDSR). (a) RIRR comprises a vision retriever that identifies the most semantically similar reference image $I_{\text{REF}}$ given a LR input $I_{\text{LR}}$. (b) RefDSR takes $I_{LR}$ and $I_{REF}$ as input. Text prompts extracted separately from $I_{LR}$ and $I_{REF}$ are concatenated as the text input for the UNet. $I_{REF}$ is encoded by a frozen VAE Encoder $\mathcal{E}^{REF}$ and processed by a LoRA-finetuned ControlNet $\mathcal{F}^{REF}_{\theta}$ to extract reference features. In the main pipeline, $I_{LR}$ passes through the frozen encoder $\mathcal{E}$, UNet $\mathcal{F}$, and decoder $\mathcal{D}$ without modification. Reference features are injected into the UNet decoder as additional conditions, replacing text cross-attention in the first three blocks.}
  \label{fig:pipeline}
  \vspace{-0.5cm}
\end{figure*}
\vspace{-0.5cm}
\section{Introduction}
Single Image Super Resolution (SISR) aims to reconstruct a high-resolution (HR) image from a single low-resolution (LR) input, striving to recover perceptually plausible and visually realistic textures. Recent advances, from CNN-based models \cite{bsrgan,realesrgan} to transformer architectures \cite{swinir,hat}, and diffusion-based generative models \cite{stablesr,seesr,supir}, have significantly improved the perceptual quality and fidelity of SISR results. Despite these improvements, SISR remains inherently ill-posed and often struggles to hallucinate fine-grained textures, particularly in regions with complex or ambiguous content. 

To address the limitations of SISR, Reference-based Super Resolution (RefSR) has emerged as a compelling alternative. By leveraging auxiliary high-quality reference images that share semantic or structural similarity with the LR input, RefSR methods \cite{c2matching, datsr, lmr} can transfer realistic textures and recover intricate visual details that SISR models alone typically fail to reconstruct. Despite their potential, existing RefSR approaches typically rely on manually curated target-reference image pairs, an assumption that severely limits their scalability and real-world applicability. In practice, users rarely have access to such perfectly paired references at inference time, and how to efficiently acquire suitable reference images remains an underexplored challenge.

In this work, we propose Retrieval-Augmented Super Resolution (RASR), a practical RefSR paradigm that eliminates manual reference pairing. Given any LR input, RASR automatically retrieves semantically relevant HR images from a reference database, enabling flexible deployment. High-quality reference databases are often naturally available or can be pre-collected in closed or category-specific domains (e.g., zoos or museums), making RASR particularly effective.  
To support research, we introduce RASR-Flickr30, the first benchmark with per-category reference galleries for open-world, dynamic retrieval.  
We also present RASRNet, a retrieval-to-generation baseline integrating a semantic retriever and a diffusion-based generator built on pretrained SISR-Diffusion backbones with semantic conditioning via ControlNet~\cite{controlnet}. By leveraging global semantic features rather than patch matching, RASRNet produces robust, perceptually faithful reconstructions under diverse degradations, producing realistic, detailed textures.




Our contributions are as follows: 1) We propose Retrieval-Augmented Super Resolution, a new paradigm that enables flexible reference retrieval and broadens RefSR applicability; 2) We present RASR-Flickr30, the first retrieval-augmented SR dataset with per-category reference databases for dynamic, context-aware retrieval; 3) We introduce RASRNet, a method that surpasses SISR baselines and demonstrates the potential of retrieval in real-world super-resolution.

\section{Related Work}
\subsection{Diffusion-based Real Image Super Resolution}

Diffusion models have recently excelled in high-quality image generation, from unconditional to text-to-image (T2I) tasks. Their generative capacity has motivated applications to super resolution. StableSR \cite{stablesr} adapts T2I with a time-aware encoder and feature wrapping, while DiffBIR \cite{diffbir} removes degradation before regeneration. PASD \cite{pasd} and SeeSR \cite{seesr} leverage semantic cues from image-to-text modules. However, diffusion models are costly due to many sampling steps. Efficient solutions such as SinSR \cite{sinsr} and OSEDiff \cite{osediff} accelerate inference via single-step or optimized sampling. Beyond efficiency, ReFIR \cite{refir} introduces a training-free framework that exploits external references to reduce hallucination.
\vspace{-0.1cm}
\subsection{Reference-based Image Super Resolution}

RefSR enhances reconstruction using high-quality reference images. Early works improved correspondence via alignment (flow, DCN \cite{dcn}) \cite{crossnet,ssen} or patch-matching \cite{srntt,ttsr,patchmatching}. $C^2$-Matching \cite{c2matching} combines both with correlation distillation, and DATSR \cite{datsr} employs Swin Transformer \cite{swintransformer}. MRefSR \cite{lmr} supports multiple references. While surpassing SISR \cite{realesrgan,swinir}, these methods often require paired references and patch matching, limiting real-world use \cite{stablesr}. Thus, efficiently retrieving suitable references and aggregating their information without precise matching remains a key challenge that this work seeks to address.

\begin{table*}[ht]
\small
	\centering
	\begin{threeparttable}
        \begin{tabular}{cccccccccc}
			\toprule
			\multirow{1}{*}{Method}&& PSNR$\uparrow$  & SSIM$\uparrow$ & LPIPS$\downarrow$ & DISTS$\downarrow$ & FID$\downarrow$ & NIQE$\downarrow$ &  MUSIQ$\uparrow$ &  CLIPIQA$\uparrow$ \cr
			\midrule
            \midrule

            BSRGAN&\multirow{3}{*}{GAN-based}&24.72&0.5952&0.4399&0.2774&147.79&4.9689&54.36&0.5851\cr
            Real-ESRGAN&&24.78&0.6078&0.4268&0.2694&124.53&4.9465&52.52&0.5373\cr
            HAT&&25.07&0.6160&0.4007&0.2804&110.63&5.3254&48.20&0.4580\cr
            \midrule
RAG-$C^2$-Matching &\multirow{3}{*}{Reference-based}&24.40 & 0.5663 &0.7004&0.3503&155.06&8.6838&17.97&0.1587\cr
			RAG-DATSR &&24.41&0.5677&0.7052&0.3500&153.50&9.1758&17.99&0.1621\cr
			RAG-MRefSR &&24.47&0.5753&0.7466&0.3395&153.25&9.9936&17.55&0.3004\cr
            \midrule
StableSR&\multirow{5}{*}{Diffusion-based}&22.88&0.4867&0.3952&0.2147&87.74&4.3289&66.30&0.7083\cr
            DiffBIR&&23.26&0.5144&0.4266&0.2236&109.57&5.0749&73.46&0.8366\cr
            PASD&&23.97&0.5773&0.4724&0.2403&98.07&5.2452&61.61&0.6130\cr
            SeeSR&&23.26&0.5515&0.3740&0.1919&72.97&4.4319&72.89&0.7987\cr
            OSEDiff&&23.29&0.5624&0.3404&0.1915&65.59&4.2009&71.35&0.7572\cr
            \midrule
		  
		\multirow{2}{*}{RASRNet (Ours)}&\multirow{2}{*}{RASR}&23.67&0.5687&0.3273&0.1743&56.83&4.1139&72.33&0.7706\cr
        &&\textit{{+0.38}}&\textit{{+0.0063}}&\textit{{-0.0131}}&\textit{{-0.0172}}&\textit{{-8.76}}&\textit{{-0.0870}}&\textit{{+0.98}}&\textit{{+0.0134}}\cr
			\bottomrule
			\bottomrule
		\end{tabular}

		\caption{Quantitative comparison with state-of-the-art methods on RASR-Flickr30 test set.}
		\label{tab:table1}
	\end{threeparttable}
    \vspace{-0.5cm}
\end{table*}

\section{Methodology}

\subsection{Task Formulation: RASR}

As detailed in the Introduction, traditional RefSR methods rely on manually curated target-reference pairs, limiting their scalability and flexibility. To overcome these limitations, we propose Retrieval-Augmented Super Resolution (RASR), a new RefSR paradigm that automatically retrieves semantically relevant references from a reference database to guide the super resolution process for LR input $I_{LR}$.
Formally, the RASR task is defined as:
\begin{equation}
\text{RASR}(I_{LR})=\mathcal{G}(I_{LR}, I_{REF})=\mathcal{G}(I_{LR},\mathcal{R}(\mathcal{DB}, I_{LR})),
\end{equation}
where $\mathcal{DB}$ denotes a  pre-curated database containing high-quality reference images, $\mathcal{R}$ is a retriever that selects the most semantically relevant reference image $I_{REF}$ from $\mathcal{DB}$ based on $I_{LR}$, $\mathcal{G}$ is a generation function that reconstructs HR output conditioned on both $I_{LR}$ and $I_{REF}$.
 \vspace{-0.1cm}
\subsection{RASR Network}
\noindent\textbf{RIRR: Real-world Image Reference Retriever}\
Unlike texture-based methods that fail on degraded LR images, semantic-based retrieval using vision encoders remains robust under real-world conditions. As shown in Fig.\ref{fig:pipeline} (a), RIRR extracts embeddings from reference images using DINOv2\cite{dinov2}, then retrieves the most similar reference for each LR input via cosine similarity matching.

\noindent\textbf{RefDSR: Reference Diffusion Super Resolution} \\
Given a LR input image \textbf{$I_{LR}$} and a reference image \textbf{$I_{REF}$} retrieved from the reference database $\mathcal{DB}$, we propose a plug-and-play generator network $\mathcal{G}_\theta$ to produce a HR output image $\hat{I_{HR}}$. This network is designed to validate the effectiveness of the reference-based super-resolution paradigm within our retrieval-augmented generation framework.

\noindent\textbf{\textit{Framework.}}
We denote the VAE encoder, VAE decoder, and the latent diffusion network of a pretrained Diffusion model for SISR task as $\mathcal{E}$, $\mathcal{D}$, and $\mathcal{F}$, respectively. To preserve the original generative capability of the backbone, we freeze all parameters of the pretrained modules and introduce a trainable ControlNet \cite{controlnet} branch, denoted as $\mathcal{F}^{REF}_{\theta}$, as shown in Fig. \ref{fig:pipeline} (b).
In detail, we introduce lightweight, trainable components -- LoRA~\cite{lora} modules and train zero-convolution layers in the DownBlocks of $\mathcal{F}^{REF}_{\theta}$.
To encode \textbf{$I_{REF}$} into the latent space, we utilize a frozen VAE encoder $\mathcal{E}^{REF}$ from a pretrained diffusion model. The encoded latent representation $z_{REF}$ of the reference image is then fed into the ControlNet branch $\mathcal{F}^{REF}_{\theta}$ as the conditioning input. To obtain an unconditioned latent representation of the reference image, we exclude text conditioning during its encoding.

The reference image provides semantic rather than structural similarity to the input, and injecting its features into shallow layers can mislead low-level reconstruction. In the vanilla ControlNet architecture, conditional features are injected into all decoder blocks, with text embeddings guiding the entire UNet via cross-attention. In contrast, our RefDSR model injects reference conditioning only into the first three decoder blocks (see Eq.~\ref{eq:fuse}), while text-based cross-attention is applied exclusively in the final decoder block. Here, $f_{\text{UNet}}^{i}$ and $f_{\text{ControlNet}}^{i}$ denote the UNet and ControlNet features at block $i$, respectively. The updated UNet feature is:
\vspace{-0.1cm}
\begin{equation}
\hat{f}_{\text{UNet}}^{i}
=
f_{\text{UNet}}^{i}
+
f_{\text{ControlNet}}^{i}, \quad i \in \{0,1,2\}.
\label{eq:fuse}
\end{equation}
\vspace{-0.1cm}
Importantly, no text-based modulation is applied in these first three layers; only the ControlNet features are fused. This design ensures that the early decoding stages focus on reconstructing structural details under semantic guidance from the reference image, while global fidelity and style are refined by text conditioning at the final stage.

The reference diffusion process can be written as:
\begin{equation}
\begin{split}
\hat{I_{HR}} &= \mathcal{G_\theta}(I_{LR}, I_{REF}, c_{text}) \\
&\triangleq \mathcal{D}(\mathcal{F} (\mathcal{E}(I_{LR}); \mathcal{F}^{REF}_{\theta}(\mathcal{E}^{c}(I_{REF})); c_{text})),
\end{split}
\end{equation}
where $c_{text}$ denotes the conditional text in the pretrained SISR Diffusion model.

\noindent\textbf{\textit{Training Objectives.}}
During training, we use an MSE loss, a perceptual LPIPS loss and a GAN loss, where the discriminator adopts a fixed DINO \cite{dino} backbone, following \cite{dinogan}. To generate sharper and more realistic details, we additionally adopt a Gram loss $\mathcal{L}_{\text{Gram}}$, defined as the L2 norm of the difference between the Gram matrices of VGG-16 features:

\begin{equation}
\mathcal{L}_{\text{Gram}}=\sum_{l \in \mathcal{L}} \lambda_l \frac{1}{C_l H_l W_l}\left\|G_l(\hat{I_{HR}})-G_l(I_{GT})\right\|_2^2,
\end{equation}

\noindent where $G_l(\cdot)$ denotes the Gram matrix computed from the feature maps at layer $l$, $\lambda_l$ is the layer weight, and $I_{GT}$ denotes the ground truth (GT) image.

The overall training loss is formulated as a weighted sum of the aforementioned terms:
\begin{equation}
\mathcal{L} = \mathcal{L}_{\text{MSE}} + \lambda_{\text{LPIPS}}\cdot\mathcal{L}_{\text{LPIPS}} + \lambda_\text{Gram}\cdot \mathcal{L}_{\text{Gram}} + \lambda_{\text{GAN}}\cdot\mathcal{L}_{\text{GAN}},
\end{equation}
where $\lambda_{\text{LPIPS}}, \lambda_\text{Gram}, \lambda_{\text{GAN}}$ are the loss weights for each term.



\begin{figure*}[ht]
	\centering
\begin{minipage}{0.93\textwidth} 
    \centering
	\begin{minipage}{0.332\textwidth}
		\centering
	\includegraphics[width=\linewidth]{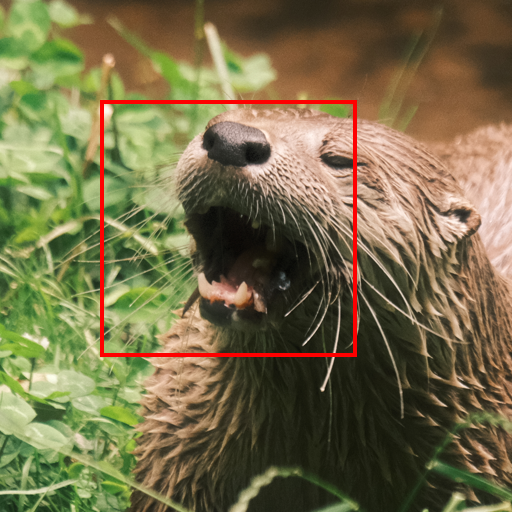}\\
        \caption*{}
	\end{minipage}
    \hspace{0.5pt}    
	\begin{minipage}{0.15\textwidth}
		\centering
		\begin{subfigure}{\linewidth}
        \centering
        \includegraphics[width=\linewidth]{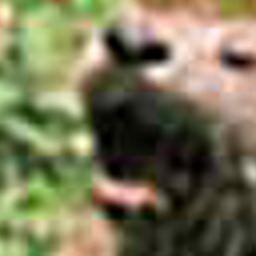}
        \caption*{Zoomed LR}
    \end{subfigure}
    \begin{subfigure}{\linewidth}
        \centering
        \includegraphics[width=\linewidth]{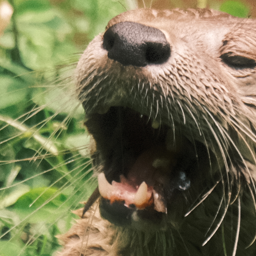}
        \caption*{Zoomed GT}
    \end{subfigure}
	\end{minipage}
    \hspace{0.5pt}
	\begin{minipage}{0.15\textwidth}
		\centering
        \begin{subfigure}{\linewidth}
        \centering
		\includegraphics[width=\linewidth]{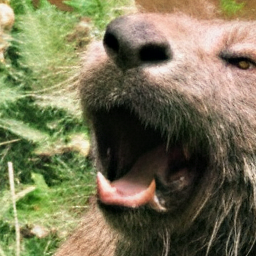}
        \caption*{StableSR}
    \end{subfigure}
        \begin{subfigure}{\linewidth}
        \centering
        \includegraphics[width=\linewidth]{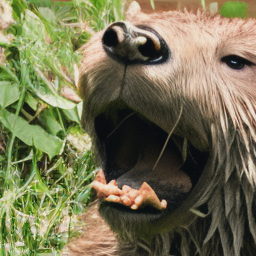}
        \caption*{SeeSR}
    \end{subfigure}
	\end{minipage}
    \hspace{0.5pt}
	\begin{minipage}{0.15\textwidth}
		\centering
        \begin{subfigure}{\linewidth}
        \centering
		\includegraphics[width=\linewidth]{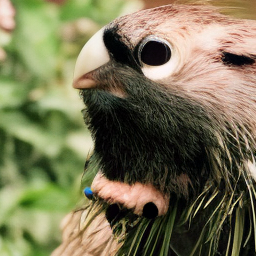} 
        \caption*{DiffBIR}
    \end{subfigure}
        \begin{subfigure}{\linewidth}
        \centering
        \includegraphics[width=\linewidth]{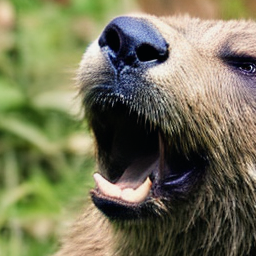}
        \caption*{OSEDiff}
    \end{subfigure}
	\end{minipage}
    \hspace{0.5pt}
    \begin{minipage}{0.15\textwidth}
		\centering
        \begin{subfigure}{\linewidth}
        \centering
		\includegraphics[width=\linewidth]{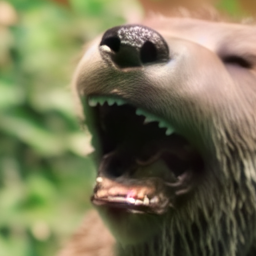}
        \caption*{PASD}
    \end{subfigure}
        \begin{subfigure}{\linewidth}
        \centering
        \includegraphics[width=\linewidth]{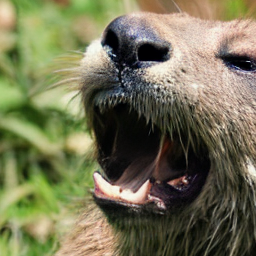}
        \caption*{RASRNet (ours)}
    \end{subfigure}
	\end{minipage}
\end{minipage}

\begin{minipage}{0.93\textwidth} 
    \centering
	\begin{minipage}{0.332\textwidth}
		\centering
	\includegraphics[width=\linewidth]{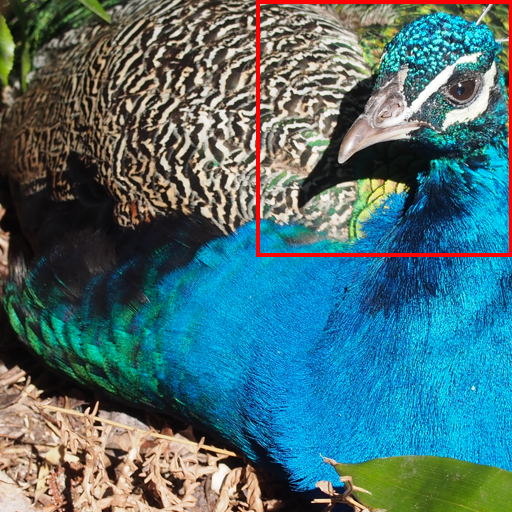}\\
        \caption*{}
	\end{minipage}
    \hspace{0.5pt}    
	\begin{minipage}{0.15\textwidth}
		\centering
		\begin{subfigure}{\linewidth}
        \centering
        \includegraphics[width=\linewidth]{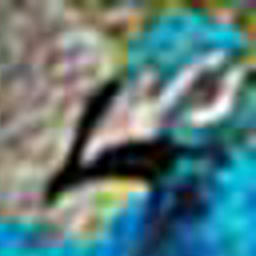}
        \caption*{Zoomed LR}
    \end{subfigure}
    \begin{subfigure}{\linewidth}
        \centering
        \includegraphics[width=\linewidth]{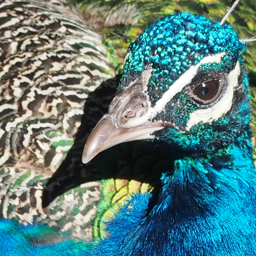}
        \caption*{Zoomed GT}
    \end{subfigure}
	\end{minipage}
    \hspace{0.5pt}
	\begin{minipage}{0.15\textwidth}
		\centering
        \begin{subfigure}{\linewidth}
        \centering
		\includegraphics[width=\linewidth]{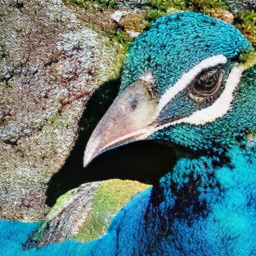}
        \caption*{StableSR}
    \end{subfigure}
        \begin{subfigure}{\linewidth}
        \centering
        \includegraphics[width=\linewidth]{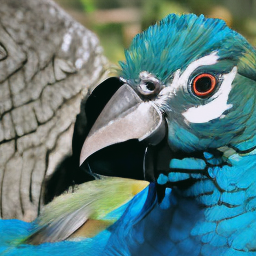}
        \caption*{SeeSR}
    \end{subfigure}
	\end{minipage}
    \hspace{0.5pt}
	\begin{minipage}{0.15\textwidth}
		\centering
        \begin{subfigure}{\linewidth}
        \centering
		\includegraphics[width=\linewidth]{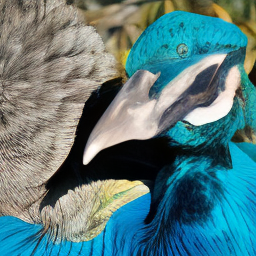} 
        \caption*{DiffBIR}
    \end{subfigure}
        \begin{subfigure}{\linewidth}
        \centering
        \includegraphics[width=\linewidth]{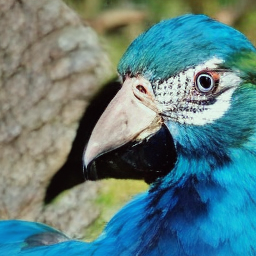}
        \caption*{OSEDiff}
    \end{subfigure}
	\end{minipage}
    \hspace{0.5pt}
    \begin{minipage}{0.15\textwidth}
		\centering
        \begin{subfigure}{\linewidth}
        \centering
		\includegraphics[width=\linewidth]{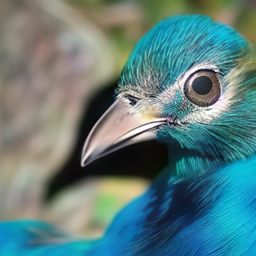}
        \caption*{PASD}
    \end{subfigure}
        \begin{subfigure}{\linewidth}
        \centering
        \includegraphics[width=\linewidth]{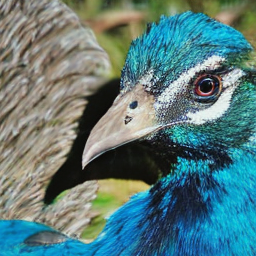}
        \caption*{RASRNet (ours)}
    \end{subfigure}
	\end{minipage}
\end{minipage}
    \caption{Qualitative comparisons of different diffusion-based SR methods.}
    \label{fig:srs_comparison}
    \vspace{-0.5cm}
\end{figure*}

\subsection{RASR-Flickr30 Dataset}

We propose RASR-Flickr30, a dataset of high-quality target and reference images collected from Flickr.com\footnote{\url{https://www.flickr.com/}}. Focusing on animals for their diversity yet intra-species consistency, the dataset reflects practical RASR scenarios. It includes 30 species with variations in pose, background, lighting, and viewpoint. For each species, 40 images are used for training, 5 for testing, and about 100 as references to support retrieval.

\section{Experiments}

\renewcommand{\arraystretch}{1.2} 

\subsection{Experiments Settings}

\noindent\textbf{Training and Testing Datasets.}
We train RefDSR on RASR-Flickr30 and the LSDIR \cite{lsdir} dataset.
For testing, we use the RASR-Flickr30 test set and the WR-SR \cite{c2matching} dataset. To generate LR and HR training and testing pairs, we adopt the degradation pipeline from Real-ESRGAN \cite{realesrgan}.

\noindent\textbf{Evaluation Metrics.}
For evaluation, we adopt both fidelity metrics (PSNR and SSIM) and perceptual metrics (LPIPS \cite{lpips}, DISTS \cite{dists}, and FID \cite{fid}), along with several no-reference metrics (NIQE \cite{niqe}, MUSIQ \cite{musiq}, and CLIPIQA \cite{clipiqa}) to comprehensively assess the performance of different methods. 


\noindent\textbf{Compared Methods.} We compare our RASRNet with state-of-the-art GAN-based SISR methods, including BSRGAN \cite{bsrgan}, Real-ESRGAN \cite{realesrgan}, and HAT \cite{hat}; RefSR methods, including $C^2$-Matching \cite{c2matching}, DATSR \cite{datsr}, and MRefSR \cite{lmr}; as well as diffusion-based methods, including StableSR \cite{stablesr}, DiffBIR \cite{diffbir}, PASD \cite{pasd}, SeeSR \cite{seesr}, and OSEDiff \cite{osediff}.

\subsection{Comparison with State-of-the-Arts}
\noindent\textbf{Quantitative Comparison.} For the retrieval-based setting, we present quantitative comparisons on the synthetic RASR-Flickr30 test set in Tab. \ref{tab:table1}. For RefSR methods, we use the reference images retrieved by our RIRR module. We have the following observations:  
(1) Compared to the baseline OSEDiff, our RASRNet achieves consistent improvements across both full-reference and no-reference metrics. Specifically, it brings a PSNR improvement of +0.38~dB and an SSIM improvement of +0.0063. For perceptual metrics, it achieves reductions of -0.0131 in LPIPS, -0.0172 in DISTS, and -8.76 in FID, indicating enhanced visual fidelity and texture realism. In terms of no-reference metrics, RASRNet further improves NIQE by -0.0870 and achieves gains of +0.98 in MUSIQ and +0.0134 in CLIPIQA.
(2) Among all evaluated methods, RASRNet attains the best results on LPIPS (0.3273), DISTS (0.1743), FID (56.83), and NIQE (4.1139), demonstrating its superiority in perceptual quality and realistic texture reconstruction.
(3) Previous RefSR methods exhibit inferior performance on images with realistic degradations compared to GAN-based SISR models. This is largely due to the reliance on patch-matching correspondence networks, which become unreliable under complex or unknown degradations. In contrast, our RefDSR model leverages a diffusion-based architecture to enable more robust and reliable reference utilization, making RefSR truly effective in the presence of realistic degradations.




\noindent\textbf{Qualitative Comparison.}
Fig. \ref{fig:srs_comparison} presents visual comparisons of different diffusion-based SR methods.
As illustrated in the first example, the input image contains an otter. StableSR fails to fully remove noise, DiffBIR does not reconstruct the correct otter structure, and PASD produces overly smooth textures lacking fine details. Although SeeSR and OSEDiff leverage degradation-aware semantic cues to generate reasonable results, they still lack realistic details, making their outputs appear less natural. In contrast, RASRNet successfully reconstructs richer fur textures while also generating a more realistic nose and mouth, resulting in a faithful and visually compelling restoration.
A similar trend is observed in the second example. StableSR incorrectly interprets the noisy background as a mountain, DiffBIR fails to generate the correct structure of the peacock, and PASD again produces overly smoothed textures. SeeSR and OSEDiff deliver reasonable but detail-lacking results. RASRNet, however, reconstructs finer head textures and neck feathers, achieving a more realistic and structurally accurate output. 

\section{Conclusion}

In this paper, we propose RASR, a retrieval-augmented paradigm for practical reference-based image restoration. It integrates semantic retrieval from a reference database to eliminate the need for curated target–reference image pairs required by current RefSR methods. To advance research in this domain, we construct RASR-Flickr30, the first benchmark with per-category reference databases for open-world retrieval. Our proposed baseline, RASRNet, combines semantic retrieval with a diffusion-based generator and demonstrates clear improvements over SISR baselines. These results underscore the potential of retrieval augmentation in bridging the gap between academic RefSR research and real-world applications.

\newpage
\bibliographystyle{IEEEtran}
\bibliography{main}

@article{osediff,
  title={One-step effective diffusion network for real-world image super-resolution},
  author={Wu, Rongyuan and Sun, Lingchen and Ma, Zhiyuan and Zhang, Lei},
  journal={Advances in Neural Information Processing Systems},
  volume={37},
  pages={92529--92553},
  year={2024}
}

@inproceedings{sinsr,
  title={Sinsr: diffusion-based image super-resolution in a single step},
  author={Wang, Yufei and Yang, Wenhan and Chen, Xinyuan and Wang, Yaohui and Guo, Lanqing and Chau, Lap-Pui and Liu, Ziwei and Qiao, Yu and Kot, Alex C and Wen, Bihan},
  booktitle={Proceedings of the IEEE/CVF conference on computer vision and pattern recognition},
  pages={25796--25805},
  year={2024}
}

@inproceedings{c2matching,
  title={Robust reference-based super-resolution via c2-matching},
  author={Jiang, Yuming and Chan, Kelvin CK and Wang, Xintao and Loy, Chen Change and Liu, Ziwei},
  booktitle={Proceedings of the IEEE/CVF Conference on Computer Vision and Pattern Recognition},
  pages={2103--2112},
  year={2021}
}

@inproceedings{datsr,
  title={Reference-based image super-resolution with deformable attention transformer},
  author={Cao, Jiezhang and Liang, Jingyun and Zhang, Kai and Li, Yawei and Zhang, Yulun and Wang, Wenguan and Gool, Luc Van},
  booktitle={European conference on computer vision},
  pages={325--342},
  year={2022},
  organization={Springer}
}

@inproceedings{lmr,
  title={LMR: a large-scale multi-reference dataset for reference-based super-resolution},
  author={Zhang, Lin and Li, Xin and He, Dongliang and Li, Fu and Ding, Errui and Zhang, Zhaoxiang},
  booktitle={Proceedings of the IEEE/CVF International Conference on Computer Vision},
  pages={13118--13127},
  year={2023}
}

@article{refir,
  title={Refir: Grounding large restoration models with retrieval augmentation},
  author={Guo, Hang and Dai, Tao and Ouyang, Zhihao and Zhang, Taolin and Zha, Yaohua and Chen, Bin and Xia, Shu-tao},
  journal={Advances in Neural Information Processing Systems},
  volume={37},
  pages={46593--46621},
  year={2024}
}

@inproceedings{seesr,
  title={Seesr: Towards semantics-aware real-world image super-resolution},
  author={Wu, Rongyuan and Yang, Tao and Sun, Lingchen and Zhang, Zhengqiang and Li, Shuai and Zhang, Lei},
  booktitle={Proceedings of the IEEE/CVF conference on computer vision and pattern recognition},
  pages={25456--25467},
  year={2024}
}

@article{fid,
  title={Gans trained by a two time-scale update rule converge to a local nash equilibrium},
  author={Heusel, Martin and Ramsauer, Hubert and Unterthiner, Thomas and Nessler, Bernhard and Hochreiter, Sepp},
  journal={Advances in neural information processing systems},
  volume={30},
  year={2017}
}

@article{niqe,
  title={A feature-enriched completely blind image quality evaluator},
  author={Zhang, Lin and Zhang, Lei and Bovik, Alan C},
  journal={IEEE Transactions on Image Processing},
  volume={24},
  number={8},
  pages={2579--2591},
  year={2015},
  publisher={IEEE}
}

@inproceedings{clipiqa,
  title={Exploring clip for assessing the look and feel of images},
  author={Wang, Jianyi and Chan, Kelvin CK and Loy, Chen Change},
  booktitle={Proceedings of the AAAI conference on artificial intelligence},
  volume={37},
  pages={2555--2563},
  year={2023}
}

@inproceedings{musiq,
  title={Musiq: Multi-scale image quality transformer},
  author={Ke, Junjie and Wang, Qifei and Wang, Yilin and Milanfar, Peyman and Yang, Feng},
  booktitle={Proceedings of the IEEE/CVF international conference on computer vision},
  pages={5148--5157},
  year={2021}
}

@inproceedings{realesrgan,
  title={Real-esrgan: Training real-world blind super-resolution with pure synthetic data},
  author={Wang, Xintao and Xie, Liangbin and Dong, Chao and Shan, Ying},
  booktitle={Proceedings of the IEEE/CVF international conference on computer vision},
  pages={1905--1914},
  year={2021}
}

@inproceedings{lpips,
  title={The unreasonable effectiveness of deep features as a perceptual metric},
  author={Zhang, Richard and Isola, Phillip and Efros, Alexei A and Shechtman, Eli and Wang, Oliver},
  booktitle={Proceedings of the IEEE conference on computer vision and pattern recognition},
  pages={586--595},
  year={2018}
}

@article{dists,
  title={Image quality assessment: Unifying structure and texture similarity},
  author={Ding, Keyan and Ma, Kede and Wang, Shiqi and Simoncelli, Eero P},
  journal={IEEE transactions on pattern analysis and machine intelligence},
  volume={44},
  number={5},
  pages={2567--2581},
  year={2020},
  publisher={IEEE}
}

@article{lora,
  title={Lora: Low-rank adaptation of large language models.},
  author={Hu, Edward J and Shen, Yelong and Wallis, Phillip and Allen-Zhu, Zeyuan and Li, Yuanzhi and Wang, Shean and Wang, Lu and Chen, Weizhu and others},
  journal={ICLR},
  volume={1},
  number={2},
  pages={3},
  year={2022}
}

@inproceedings{hat,
  title={Activating more pixels in image super-resolution transformer},
  author={Chen, Xiangyu and Wang, Xintao and Zhou, Jiantao and Qiao, Yu and Dong, Chao},
  booktitle={Proceedings of the IEEE/CVF conference on computer vision and pattern recognition},
  pages={22367--22377},
  year={2023}
}

@article{stablesr,
  title={Exploiting diffusion prior for real-world image super-resolution},
  author={Wang, Jianyi and Yue, Zongsheng and Zhou, Shangchen and Chan, Kelvin CK and Loy, Chen Change},
  journal={International Journal of Computer Vision},
  volume={132},
  number={12},
  pages={5929--5949},
  year={2024},
  publisher={Springer}
}

@inproceedings{srntt,
  title={Image super-resolution by neural texture transfer},
  author={Zhang, Zhifei and Wang, Zhaowen and Lin, Zhe and Qi, Hairong},
  booktitle={Proceedings of the IEEE/CVF conference on computer vision and pattern recognition},
  pages={7982--7991},
  year={2019}
}

@inproceedings{ttsr,
  title={Learning texture transformer network for image super-resolution},
  author={Yang, Fuzhi and Yang, Huan and Fu, Jianlong and Lu, Hongtao and Guo, Baining},
  booktitle={Proceedings of the IEEE/CVF conference on computer vision and pattern recognition},
  pages={5791--5800},
  year={2020}
}

@article{dinov2,
  title={Dinov2: Learning robust visual features without supervision},
  author={Oquab, Maxime and Darcet, Timoth{\'e}e and Moutakanni, Th{\'e}o and Vo, Huy and Szafraniec, Marc and Khalidov, Vasil and Fernandez, Pierre and Haziza, Daniel and Massa, Francisco and El-Nouby, Alaaeldin and others},
  journal={arXiv preprint arXiv:2304.07193},
  year={2023}
}

@inproceedings{controlnet,
  title={Adding conditional control to text-to-image diffusion models},
  author={Zhang, Lvmin and Rao, Anyi and Agrawala, Maneesh},
  booktitle={Proceedings of the IEEE/CVF international conference on computer vision},
  pages={3836--3847},
  year={2023}
}

@inproceedings{bsrgan,
  title={Designing a practical degradation model for deep blind image super-resolution},
  author={Zhang, Kai and Liang, Jingyun and Van Gool, Luc and Timofte, Radu},
  booktitle={Proceedings of the IEEE/CVF international conference on computer vision},
  pages={4791--4800},
  year={2021}
}

@inproceedings{pasd,
  title={Pixel-aware stable diffusion for realistic image super-resolution and personalized stylization},
  author={Yang, Tao and Wu, Rongyuan and Ren, Peiran and Xie, Xuansong and Zhang, Lei},
  booktitle={European Conference on Computer Vision},
  pages={74--91},
  year={2024},
  organization={Springer}
}

@inproceedings{diffbir,
  title={Diffbir: Toward blind image restoration with generative diffusion prior},
  author={Lin, Xinqi and He, Jingwen and Chen, Ziyan and Lyu, Zhaoyang and Dai, Bo and Yu, Fanghua and Qiao, Yu and Ouyang, Wanli and Dong, Chao},
  booktitle={European Conference on Computer Vision},
  pages={430--448},
  year={2024},
  organization={Springer}
}

@inproceedings{swinir,
  title={Swinir: Image restoration using swin transformer},
  author={Liang, Jingyun and Cao, Jiezhang and Sun, Guolei and Zhang, Kai and Van Gool, Luc and Timofte, Radu},
  booktitle={Proceedings of the IEEE/CVF international conference on computer vision},
  pages={1833--1844},
  year={2021}
}

@inproceedings{swintransformer,
  title={Swin transformer: Hierarchical vision transformer using shifted windows},
  author={Liu, Ze and Lin, Yutong and Cao, Yue and Hu, Han and Wei, Yixuan and Zhang, Zheng and Lin, Stephen and Guo, Baining},
  booktitle={Proceedings of the IEEE/CVF international conference on computer vision},
  pages={10012--10022},
  year={2021}
}

@inproceedings{dcn,
  title={Deformable convolutional networks},
  author={Dai, Jifeng and Qi, Haozhi and Xiong, Yuwen and Li, Yi and Zhang, Guodong and Hu, Han and Wei, Yichen},
  booktitle={Proceedings of the IEEE international conference on computer vision},
  pages={764--773},
  year={2017}
}

@inproceedings{ssen,
  title={Robust reference-based super-resolution with similarity-aware deformable convolution},
  author={Shim, Gyumin and Park, Jinsun and Kweon, In So},
  booktitle={Proceedings of the IEEE/CVF conference on computer vision and pattern recognition},
  pages={8425--8434},
  year={2020}
}

@inproceedings{crossnet,
  title={Crossnet: An end-to-end reference-based super resolution network using cross-scale warping},
  author={Zheng, Haitian and Ji, Mengqi and Wang, Haoqian and Liu, Yebin and Fang, Lu},
  booktitle={Proceedings of the European conference on computer vision (ECCV)},
  pages={88--104},
  year={2018}
}

@article{patchmatching,
  title={PatchMatch: A randomized correspondence algorithm for structural image editing},
  author={Barnes, Connelly and Shechtman, Eli and Finkelstein, Adam and Goldman, Dan B},
  journal={ACM Trans. Graph.},
  volume={28},
  number={3},
  pages={24},
  year={2009}
}

@inproceedings{lsdir,
  title={Lsdir: A large scale dataset for image restoration},
  author={Li, Yawei and Zhang, Kai and Liang, Jingyun and Cao, Jiezhang and Liu, Ce and Gong, Rui and Zhang, Yulun and Tang, Hao and Liu, Yun and Demandolx, Denis and others},
  booktitle={Proceedings of the IEEE/CVF Conference on Computer Vision and Pattern Recognition},
  pages={1775--1787},
  year={2023}
}

@inproceedings{dinogan,
  title={Ensembling off-the-shelf models for gan training},
  author={Kumari, Nupur and Zhang, Richard and Shechtman, Eli and Zhu, Jun-Yan},
  booktitle={Proceedings of the IEEE/CVF conference on computer vision and pattern recognition},
  pages={10651--10662},
  year={2022}
}

@inproceedings{dino,
  title={Emerging properties in self-supervised vision transformers},
  author={Caron, Mathilde and Touvron, Hugo and Misra, Ishan and J{\'e}gou, Herv{\'e} and Mairal, Julien and Bojanowski, Piotr and Joulin, Armand},
  booktitle={Proceedings of the IEEE/CVF international conference on computer vision},
  pages={9650--9660},
  year={2021}
}

@inproceedings{supir,
  title={Scaling up to excellence: Practicing model scaling for photo-realistic image restoration in the wild},
  author={Yu, Fanghua and Gu, Jinjin and Li, Zheyuan and Hu, Jinfan and Kong, Xiangtao and Wang, Xintao and He, Jingwen and Qiao, Yu and Dong, Chao},
  booktitle={Proceedings of the IEEE/CVF conference on computer vision and pattern recognition},
  pages={25669--25680},
  year={2024}
}

\end{document}